\documentclass[letterpaper, 10pt, conference]{ieeeconf}

\IEEEoverridecommandlockouts                              % This command is only needed if 
\usepackage{graphics} % for pdf, bitmapped graphics files
\usepackage{epsfig} % for postscript graphics files
\usepackage{mathptmx} % assumes new font selection scheme installed
\usepackage{times} % assumes new font selection scheme installed
\usepackage{amsmath} % assumes amsmath package installed
\usepackage{amssymb}  % assumes amsmath package installed
\usepackage{graphicx}
\usepackage[font=footnotesize]{caption}
\usepackage[font=footnotesize]{subcaption}
\usepackage{dblfloatfix}
\usepackage[ruled, linesnumbered]{algorithm2e}
\usepackage[sortcites,backend=bibtex,citestyle=numeric,bibstyle=ieee,sorting=none,natbib=true,maxcitenames=1,mincitenames=1]{biblatex}

\usepackage{xcolor}

\title{\LARGE \bf Ctrl-Z: Recovering from Instability in Reinforcement Learning}

\author{
  Vibhavari Dasagi$^{1}$, Jake Bruce$^{1}$, Thierry Peynot$^{1}$, J\"urgen Leitner$^{1}$
\vspace{-2mm}
\thanks{This research was supported by the Australian Research Council Centre of Excellence for Robotic Vision (project number CE140100016).}%
\thanks{$^{1}$ Australian Centre for Robotic Vision (ACRV), Queensland University of Technology (QUT), Brisbane, Australia}%
\thanks{
Contact: {\tt\small vibhavari.dasagi@hdr.qut.edu.au}}%
}

\begin{document}

\maketitle
\thispagestyle{empty}
\pagestyle{empty}

%===============================================================================

\begin{abstract}

When learning behavior, training data is often generated by the learner itself; this can result in unstable training dynamics, and this problem has particularly important applications in safety-sensitive real-world control tasks such as robotics. In this work, 
we propose a principled and model-agnostic approach to mitigate
the issue of unstable learning dynamics by maintaining a history of a reinforcement learning agent over
the course of training, and reverting to the parameters of a previous agent whenever
performance significantly decreases.
We develop techniques for evaluating this performance through statistical hypothesis testing of continued improvement, and evaluate them on a standard suite of challenging benchmark tasks involving continuous
control of simulated robots.
We show improvements over state-of-the-art reinforcement learning algorithms in performance and robustness to hyperparameters, outperforming DDPG in 5 out of 6 evaluation environments and showing no decrease in performance with TD3, which is known to be relatively stable. In this way, our approach takes an important step towards increasing data efficiency and stability in training for real-world robotic applications. 

\end{abstract}

%===============================================================================

\section{Introduction}

Online behavior learning, typically in the form of deep reinforcement learning (RL), has
demonstrated significant successes in recent years~\cite{mnih2015human, mnih2016asynchronous, silver2016mastering, lillicrap2015continuous}. In these algorithms, a learning agent begins with minimal knowledge of the
domain, and discovers behavior through trial and error to optimize a task-dependent reward function.
An agent of this form implements a recursive procedure in which it generates a training dataset for itself
to learn from, bootstrapping from a blank slate to useful behavior effectively through self-supervision.

A potential issue in these recursive learning systems is the importance of monotonic improvement, and the
consequences of performance degradation during training. As long as the system is improving, it continues to
provide itself with useful and relevant training data. However, if performance decreases, due to
errors in function approximation~\cite{thrun1993issues, anschel2017averaged, fujimoto2018addressing},
suboptimal hyperparameters~\cite{jaderberg2017population, haarnoja2018soft},
or poor exploration~\cite{thrun1992efficient, osband2016deep, fortunato2017noisy},
the data being generated may be insufficient for the agent to swiftly recover. This
can result in unstable learning dynamics that dramatically reduce data efficiency,
and has implications for safety-sensitive domains in which sudden performance decreases can be dangerous.

The issues of data efficiency and safety have particular impact on the applicability of RL in robotics.
Robot training data can be expensive in terms of power costs, hardware failure and human supervision,
and unstable learning dynamics exacerbate all of these expenses.
Although the training time can be reduced through parallelization and extensive data
re-use~\cite{levine2018learning, kalashnikov2018qt},
data efficiency is always an important metric to improve, and recovering from performance degradation as early
as possible is a priority whenever real-world hardware is involved.

\begin{figure}[tbp]
    \centering
    \includegraphics[width=\columnwidth]{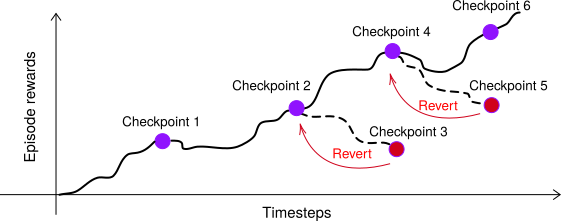}
    \caption{Our proposed method periodically evaluates the current policy of a reinforcement learning agent against previous checkpoints, calculating the probability of continued policy improvement based on the evaluation rewards. If the probability is lower than an empirically estimated threshold, the network parameters are reverted to the previously best performing policy, avoiding further performance degradation.}
    \label{fig:rollback}
\end{figure}

In this work, we address the issue of recovering from failures during training, by introducing a principled
mechanism for rolling back parameter updates as soon as they result in a significant decrease in performance (see Figure~\ref{fig:rollback}).
We develop and evaluate a family of related techniques that dynamically test the hypothesis that the
agent is continuing to improve. Our approach is completely agnostic to choices of model architecture,
RL algorithm, task domain, and specific details of the reward function such as density and magnitude.
We demonstrate performance improvements over state-of-the-art deep RL algorithms on a standard benchmark suite of simulated
continuous robot control tasks, including robustness to variations in nominal values of hyperparameters and excessively high learning
rates that cause unaugmented algorithms to fail. We also show that the threshold parameter we introduce
is not sensitive and has a wide range of high-performing values. Our approach also has very low computational
overhead, making it an attractive candidate to improve stability of a wide variety of online behavior learning 
approaches with little cost.

%===============================================================================

\section{Related Work}

When learning behavior online, tasks are typically formulated as Markov Decision Processes (MDPs), with a particular environment defined by a set of states, actions, and a transition and reward function. The objective of the behavior policy is to map states to actions in order to maximize the sum of discounted rewards, and RL approaches this optimization problem by learning from self-supervised data. Training using self-generated data relies heavily on the quality of the data gathering policy, such that
exploring a poor direction in parameter space can result in low-quality data insufficient for the learning
agent to quickly recover.
In this section, we discuss literature related to the issues of stability and safety in online
learning.

Policy degradation has many potential causes. Some sources of instability include value function overestimation, which can be mitigated by using multiple value functions, and the high variance of the REINFORCE gradient estimator, which is often addressed by subtracting a learned baseline \cite{mnih2015human, van2016deep, anschel2017averaged, wang2015dueling, mnih2016asynchronous, fujimoto2018addressing}. These techniques improve stability in the optimization procedure itself, but do not attempt to explicitly recover from training failures. Trust region algorithms such as TRPO~\cite{schulman2015trust} and PPO~\cite{schulman2017proximal} implement a constrained policy update in order to ensure stability and monotonic policy improvement. While constraining the magnitude of policy changes can successfully prevent large negative performance changes, it can also limit large positive improvements to the policy. In contrast, because we do not attempt to constrain the optimization procedure, our approach allows for arbitrarily large improvements to the policy while enabling recovery from catastrophic failures.

In the multi-task setting, learning different tasks sequentially without replay can result in a form of instability known as catastrophic forgetting. This phenomenon has been mitigated to some degree with various forms of distillation and network consolidation \cite{rusu2015policy, rusu2016progressive, kirkpatrick2017overcoming, teh2017distral, schwarz2018progress}. These approaches are intended to address the multi-task setting, in which the statistics of the environment (in particular, the reward function) change during the course of learning, and often assume known task boundaries. In this work, we are considering the single-task setting in which the goal is not to incorporate multiple behaviors into one policy, but to quickly recover when the single policy exhibits a significant performance decrease.

Safety is an important topic in robotics due to the danger and cost of real-world hardware, and safe learning has been addressed in contexts such as collision avoidance~\cite{kahn2017uncertainty, huang2019learning, fan2018fully}. These methods explicitly define the behavior to be avoided and are useful in environments where safety boundaries on behavior can be specified in advance, typically from cues such as force sensing.

In the domain of visual navigation, intra-episode \textit{checkpoints} have been used to enable agents to recover to earlier states mid-behavior~\cite{ma2019regretful}, which is similar in spirit to our proposal, but operating on the scale of behavior; in our work, recovery is performed on the scale of the learning procedure, spanning many episodes. Checkpoints and recovery have also been studied in the context of fault-tolerant computing systems and time warp protocols \cite{wang2007optimizing, okamura2004dynamic}, with RL used to learn dynamic checkpointing and recovery schemes.

Our approach derives from motivations similar to \textit{early stopping} in supervised learning, in which overfitting can be mitigated by monitoring validation error over the course of training, and stopping when improvement ceases~\cite{morgan1990generalization}. Although effective in the non-recursive setting of static datasets and supervised learning, the idea does not straightforwardly apply here, as RL datasets are non-stationary (due to being generated by agents that are constantly changing). In both cases however, the goal is to make sure we are always deploying the best version of our model.

To this end, we explicitly monitor policy improvement and decide when to revert, by keeping a history of previous parameters and their performance~\cite{schmidhuber1997shifting}. Population-Based Training (PBT)~\cite{jaderberg2017population} is a distributed version of this idea, evaluating agents being trained with slightly varying hyperparameters and copying high-performing parameters to low-performing individuals in the population. This makes PBT highly effective in stochastic optimization of hyperparameters, although as is common with evolutionary approaches, it comes at significant computational cost.

We propose to mitigate stability issues during the training of recursive learning methods, by introducing a principled
mechanism for rolling back parameter updates as soon as they result in a significant decrease in performance.
Our approach is agnostic to choices of model architecture,
RL algorithm, task domain, and we provide metrics that are robust to variations in reward function such as density and magnitude, and has very low computational
overhead. 
%===============================================================================

\section{Method}
\label{sec:method}

\begin{algorithm}[tbp]
%\SetAlgoNoLine%
%\KwResult{Write here the result }
 Initialize actor network $\pi_{\phi}$, critic networks $Q_{\theta_{1}}$, $Q_{\theta_{2}}$ with random parameters $\phi$, $\theta_{1}$, $\theta_{2}$\;
 Initialize target networks $\phi' \leftarrow \phi$, $\theta'_{1} \leftarrow \theta_{1}$, $\theta'_{2} \leftarrow \theta_{2}$\;
 Initialize replay buffer $\mathcal{B}$, episode number $n = 0$\;
 Randomly perturb hyperparameters\; %uniformly?
 \For{$t=1$ \KwTo $T$}{
  Select action with exploration noise $a \sim \pi(s) + \epsilon$\ and observe reward $r$ and new state $s'$\;
  Store transition tuple $(s, a, r, s')$ in $\mathcal{B}$\;
  Update $n = n + 1$ if episode done\;
  Sample mini-batch of $N$ transitions $(s, a, r, s')$ from $\mathcal{B}$ and update $\phi$, $\phi'$, $\theta_{1}$, $\theta'_{1}$, $\theta_{2}$, $\theta'_{2}$ according to standard TD3/DDPG\;
  \If{$n$ mod $N$}{
   Evaluate $\pi_{\phi}$ by repeating steps 6-7 for M episodes and observe episode rewards $R = [R_{1}, R_{2}, ... R_{M}]$\;
   Calculate $\rho = [\rho_{1}, \rho_{2}, ...]$ according to Eqn. \ref{eqn:rho} for current policy compared against each previous policy\;
   \If{$\text{min}(\rho) < \rho_{threshold}$}{
   Revert to previous policy corresponding to $\text{min}(\rho)$\;
   }
  }
 }
 \caption{Ctrl-Z}
 \label{alg:ctrlz}
\end{algorithm}

\begin{figure*}[tbp]
    \centering
    \includegraphics[width=\textwidth]{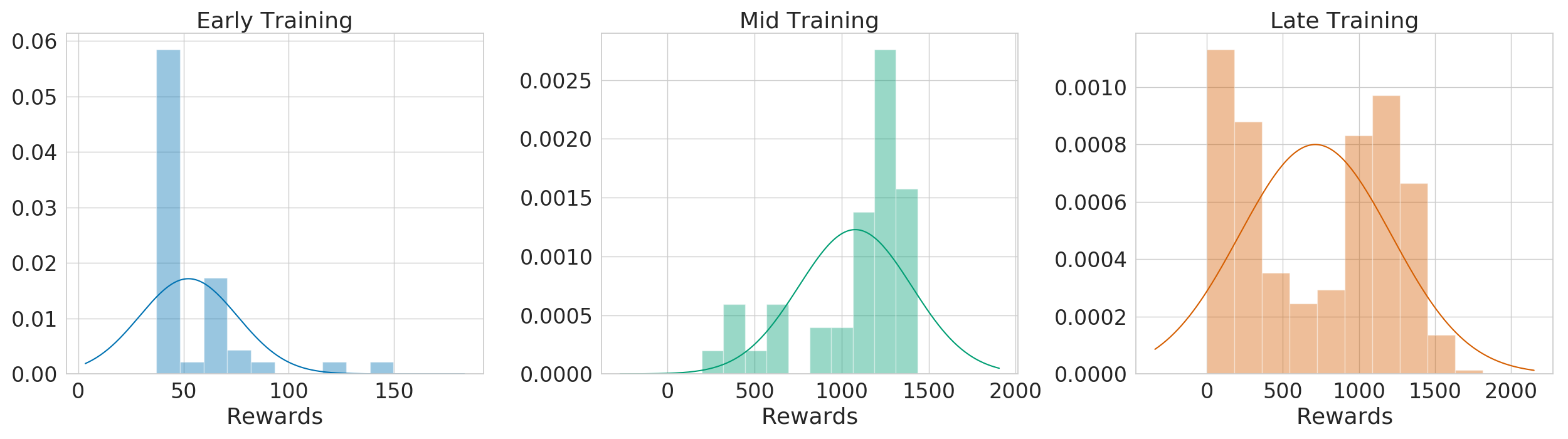}
    \caption{Examples of histograms, shown with their best corresponding Gaussian fit, over the course of training on the Hopper-v1 environment}
    \label{fig:rewardhist}
\end{figure*}

We propose a principled method for quickly detecting significant decreases in
policy performance, and propose to avoid catastrophic failure by reverting to the best performing policy. We frame the problem as a continual hypothesis testing procedure in which we are always
attempting to reject the hypothesis stating that our agent is continuing to improve. If at any 
point during the course of training we are able to reject this hypothesis, then we restore the policy
to the best performing previous version.

Our approach works as follows: After every $N$ episodes of training, we freeze the weights of the neural networks and evaluate the agent on that task for $M$ episodes, saving weights of the policy and the episodic rewards it achieved during evaluation. We then test the hypothesis that the agent is continuing to improve, by comparing the current policy performance to the previous checkpoints, thereby determining whether to revert to the last best policy (summarized in Algorithm~\ref{alg:ctrlz}). 

The choice of metric for determining when the agent should revert to a previous checkpoint is an important factor in ensuring continual performance improvement. A variety of metrics are possible, such as
comparing the distribution of the episodic rewards based on a na\"ive summary statistic such as the mean episode reward; fitting a standard distribution for each set of evaluations and analytically calculating the probability of each policy being superior; or calculating the empirical probability of superiority directly by non-parametric comparison of individual samples.

Analytical approaches based on fitting a standard distribution carry more information than simple scalar comparisons, because they can take the variance of the data into account. In general, the probability of any distribution being stochastically greater than another can be formulated as:

\begin{equation*}
    P(R_{curr} > R_{prev}) = \int_{-\infty}^{\infty}p_{curr}(x)(1 - \text{CDF}_{prev}(x))\textit{dx}
\end{equation*}

where $p_{curr}(x)$ is the marginal probability under the current distribution and $\text{CDF}_{prev}(x)$ is the cumulative distribution function of the previous distribution. Gaussian approximations are common, and the analytical probability of the current distribution of rewards being stochastically greater than the previous distribution is given as:

\begin{gather*}
    P(R_{curr} > R_{prev}) = 1 - \Phi\Big(-\dfrac{\mu}{\sigma}\Big)\\ 
    \mu = \mu_{R_{curr}} - \mu_{R_{prev}}\\  
    \sigma^{2} = \sigma_{R_{curr}}^{2} + \sigma_{R_{prev}}^{2} 
\end{gather*}

where $R_{curr}$ and $R_{prev}$ are the current and previous reward distributions defined by ($\mu_{R_{curr}}$,$\sigma_{R_{curr}}$) and ($\mu_{R_{prev}}$,$\sigma_{R_{prev}}$) respectively, the required distribution is a Gaussian defined by ($\mu$, $\sigma$), and $\Phi$ is the error function that defines the cumulative distribution function of that Gaussian. This approach is effective if the reward distributions are known \textit{a priori} to be approximately Gaussian. 

To evaluate whether our data is sufficiently Gaussian to use this approach, we conducted preliminary experiments and observed the distributions of total episode rewards over several episodes for our test environments. As seen in Figure \ref{fig:rewardhist}, we noticed that a Gaussian is not always a good fit to the data. In fact, fitting any typical standard distribution in cases like this would result in an inaccurate representation of the data due to heavy irregularity. Moreover, the distribution changes significantly over the course of training, invalidating the use of a single standard family of distributions. This emphasizes the importance of understanding the empirical distribution of the rewards before using an analytical comparison based on distribution fitting.

In practice, prior knowledge of the reward distribution over the course of training is not always available. In these situations, a non-parametric hypothesis test may be appropriate. We propose a metric based on the $\rho$ statistic of the Mann-Whitney U-test \cite{mcknight2010mann}, a well-established non-parametric hypothesis test in statistical analysis. The $\rho$ statistic is obtained by performing a pairwise comparison of samples to calculate how many samples from one distribution are greater than samples from the other, and dividing the result by the maximum of that value, which is the total number of comparisons:

\begin{equation}
    \rho = P(R_{curr} > R_{prev}) = \frac{1}{M^2} \sum_{i=1}^{M}\sum_{j=1}^{M}\textbf{1}(R_{curr}^{(i)} > r_{prev}^{(j)})
    \label{eqn:rho}
\end{equation}

where $R_{curr}^{(i)}$ and $R_{prev}^{(j)}$ are episodic rewards from the evaluation set for the current and previous distributions respectively, and $\textbf{1}(\cdot)$ is the indicator function that returns $1$ if and only if its argument is true. Intuitively, counting the total number of samples from one distribution that are greater than the other provides approximate distribution comparison without explicit definition of the distribution. As it is concerned only with ordinal rank, the Mann-Whitney U-test is also independent of the magnitude of the rewards, making it robust to outliers in the reward function. In both the analytical and empirical case we compare whether the test statistic is lower than a pre-defined significance threshold, in which case we reject the hypothesis of continual improvement. See Section~\ref{sec:experiments} for sensitivity analysis and selection of this significance threshold.

The evaluation tests are conducted across all checkpoints in history, and in case of a rejection of the hypothesis of continual improvement, the policy reverts to the best previous parameter set, defined as the checkpoint with the widest comparison margin against the current policy. Note that storing these checkpoints incurs only a minor cost: with some rare exceptions, complete parameters for most contemporary networks occupy only a few megabytes of memory. Where storage is a concern however, storing only a small number of the best previous networks instead of the entire history should improve resource usage while having little impact on the effectiveness of our method.

In summary, we propose an approach to reduce instability in deep RL by periodically evaluating the probability of continued policy improvement using informative methods such as distribution comparison and the Mann-Whitney U-test to attempt to reject the hypothesis of continual improvement, in which case we revert to the previous best policy. In this way, the agent has the potential to avoid catastrophic failure that would otherwise prevent the gathering of sufficiently high-quality data to recover.

%===============================================================================

\section{Experimental Results}
\label{sec:experiments}

\begin{figure}[tbp]
    \centering
    \includegraphics[width=\linewidth]{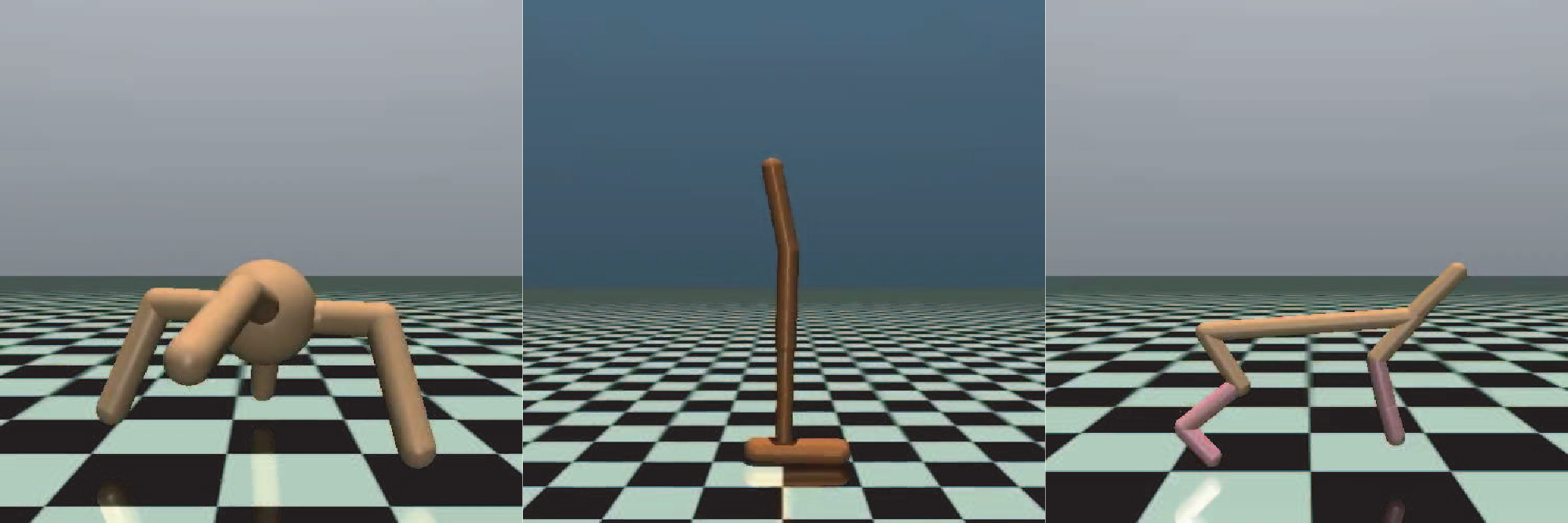}
    \caption{Examples of the standard OpenAI Gym MuJoCo environments (Left to right: Ant-v1, Hopper-v1, HalfCheetah-v1)}
    \label{fig:mujocoenvs}
\end{figure}

We evaluate our approach on a standard benchmark suite of six OpenAI Gym Mujoco environments for continuous control of simulated robots \cite{brockman2016openai, todorov2012mujoco}: HalfCheetah-v1, Ant-v1, Reacher-v1, Hopper-v1, Walker2d-v1 and Humanoid-v1 (see Figure \ref{fig:mujocoenvs}). The desired behavior in these environments is to maximize forward velocity, except for Reacher-v1 which is a target reaching task. Some of the environments also have an inherent intermediate task of balancing to be achieved before pursuing the real objective, increasing task complexity. We use state-of-the-art off-policy algorithms for continuous control, TD3 and the modified version of DDPG described in \cite{fujimoto2018addressing}, as the base learning algorithms, following the architecture and training details reported.

We alternate between training the agent for 30 episodes and evaluating for 20 episodes. The decision to revert to a previous checkpoint is based on the episode rewards achieved during the evaluation phase, as described in Section~\ref{sec:method}. For the algorithms augmented with our method, experience obtained during the evaluation stage is added to the replay buffer in order to not waste experience. To compensate for the batch of added experience, the networks are trained twice as often for every timestep spent in evaluation, to maintain the standard ratio of one training step per environment step.

In contrast to \cite{fujimoto2018addressing}, in which the agent is evaluated for 10 episodes every 5000 timesteps with no exploration noise, we report the training performance to demonstrate the effect of our method on the actual learning. Additionally, we use the RMSProp optimizer \cite{tieleman2012lecture} to isolate the effectiveness of our approach without the influence of momentum. The results are reported over 1 million timesteps, including evaluation steps, for each task, and shading represents one half standard deviation.

\subsection{Threshold Sensitivity Analysis}

We first perform a sensitivity analysis of our proposed approach to the significance threshold for rejecting the hypothesis of continual improvement. To evaluate the sensitivity of our approach to this threshold specifically for recovering from instability, we use a simple baseline-free REINFORCE gradient estimator, known to have high variance~\cite{williams1992simple}, on the standard formulation of the cartpole task with continuous actions in the Mujoco simulator; the architecture used in TD3 and DDPG but with only 32 hidden units per layer; a standard discount factor of 0.95; and we train for the same duration and use the same learning rate perturbations as the experiments in Section~\ref{sec:highlr}.

\begin{figure}[tbp]
    \centering
    \includegraphics[width=\linewidth]{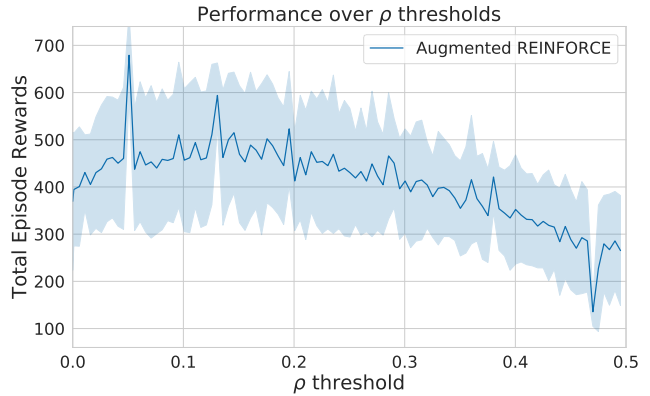}
    \caption{Reward as a function of significance threshold. Note that there is a wide range of high-performing choices, showing that our approach is not sensitive and requires no precise tuning of this hyperparameter.}
    \label{fig:thresh}
\end{figure}

We evaluate the Mann-Whitney variation of our technique with a variety of thresholds between 0 and 0.5, and we record performance over 50 seeds each. As shown in Figure \ref{fig:thresh}, our method shows insensitivity to the precise threshold value, with highest rewards achieved in the region between 0.05 and 0.2. We therefore use a value of 0.1 for the remainder of the experiments.

\subsection{High Learning Rate}
\label{sec:highlr}
A common cause of catastrophically large changes in a policy is an excessively high learning rate.
To evaluate our approach against the instability caused by this particular factor, we use DDPG to learn the tasks with a learning rate of 0.002, which is 200\% of the recommended value of 0.001 used in \cite{fujimoto2018addressing}. The experiment primarily evaluates the Mann-Whitney variant of our approach, with the comparison of mean episode rewards as an augmented baseline.

\begin{figure}[tbp]
    \centering
    \begin{subfigure}{\columnwidth}
        \centering
        \includegraphics[width=\textwidth]{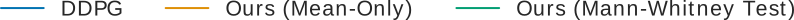}
    \end{subfigure}
    \vfill
    \begin{subfigure}{0.49\columnwidth}
        \centering
        \includegraphics[width=\textwidth]{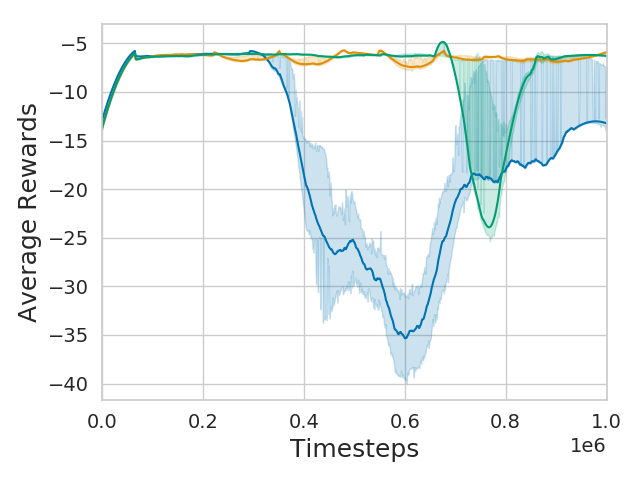}
        \caption{Reacher-v1}
    \end{subfigure}
    \hfill
    \begin{subfigure}{0.49\columnwidth}
        \centering
        \includegraphics[width=\textwidth]{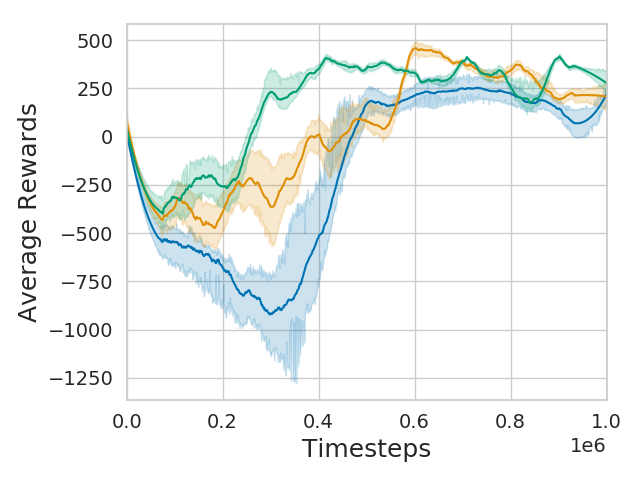}
        \caption{Ant-v1}
    \end{subfigure}
    \caption{Sample performance of DDPG on tasks with a high learning rate of 0.002}
    \label{fig:highlr}
\end{figure}

Figure \ref{fig:highlr} demonstrates that our method helps the agent recover from degenerate policies. Compared to the unaugmented approaches which take a long time to recover from performance degradation, our method detects such drops and swiftly restores performance. A higher learning rate generally results in quicker gradient descent, and so faster policy learning. However, the resulting large gradient steps may prevent the policy from settling into a minimum, and may even cause catastrophically large updates that result in policy failure. Our method is capable of recovering from these events by reverting to the previous best policy.

\subsection{Randomized Hyperparameters}

One common source of instability is suboptimal hyperparameters, and precise hyperparameter values
are often very difficult to tune. DDPG is known to be sensitive to hyperparameter values while TD3 allows for slight deviations.
In these experiments, we augment TD3 and DDPG with our method and evaluate their robustness to instability due to variations in hyperparameters by randomly perturbing all hyperparameters between 50\% and 150\% of their original values as given in \cite{fujimoto2018addressing}. These variations are greater than the range used (20\% to 120\%) in \cite{jaderberg2017population}, with far more deviation from the norm. The perturbed hyperparameters were kept constant across the different methods for each seed and the results reported across all the hyperparameter values.

\begin{figure*}[tbp]
    \centering
    \begin{subfigure}{\linewidth}
        \centering
        \includegraphics[width=0.8\linewidth]{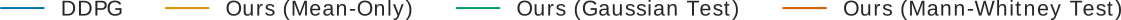}
    \end{subfigure}
    \vfill
    \begin{subfigure}{0.3\linewidth}
        \centering
        \includegraphics[width=\linewidth]{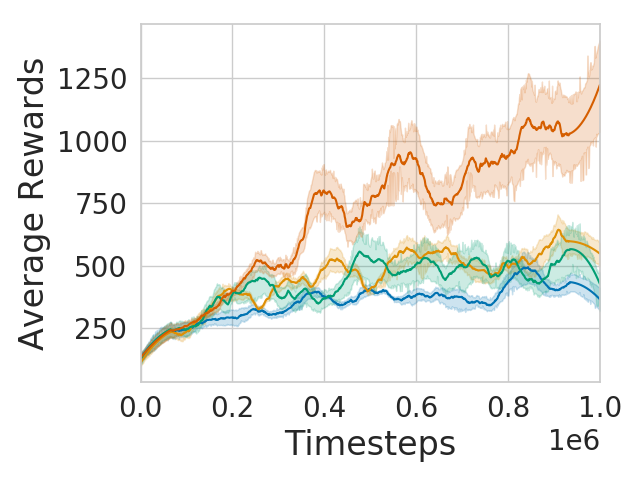}
        \caption{Walker2d-v1}
    \end{subfigure}
    \hfill
    \begin{subfigure}{0.3\linewidth}
        \centering
        \includegraphics[width=\linewidth]{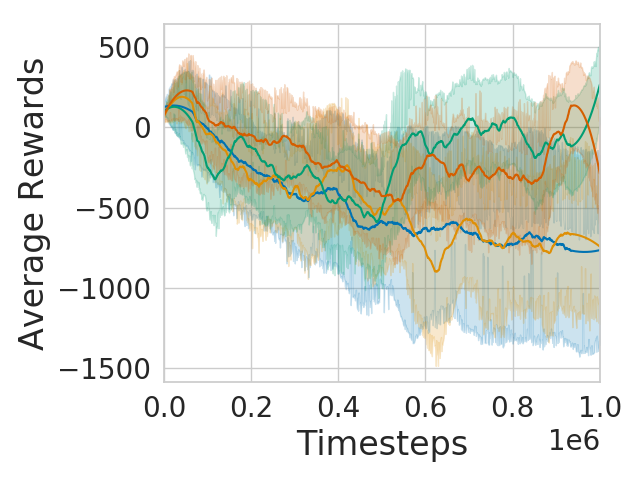}
        \caption{Ant-v1}
    \end{subfigure}
    \hfill
    \begin{subfigure}{0.3\linewidth}
        \centering
        \includegraphics[width=\linewidth]{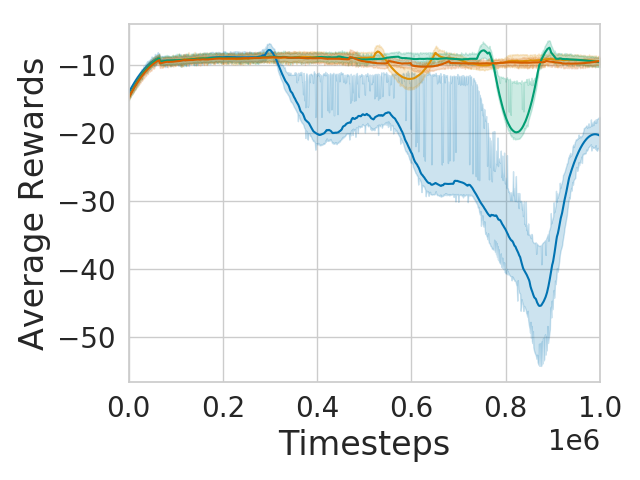}
        \caption{Reacher-v1}
    \end{subfigure}
    \caption{Sample performance curves of DDPG in the presence of hyperparameter randomization (see Fig.~\ref{fig:barplots} for full results)}
    \label{fig:randomperturbDDPG}
\end{figure*}

\begin{figure*}[tbp]
    \centering
    \begin{subfigure}{\linewidth}
        \centering
        \includegraphics[width=0.78\linewidth]{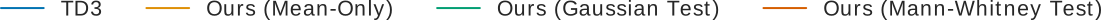}
    \end{subfigure}
    \begin{subfigure}{0.3\linewidth}
        \centering
        \includegraphics[width=\linewidth]{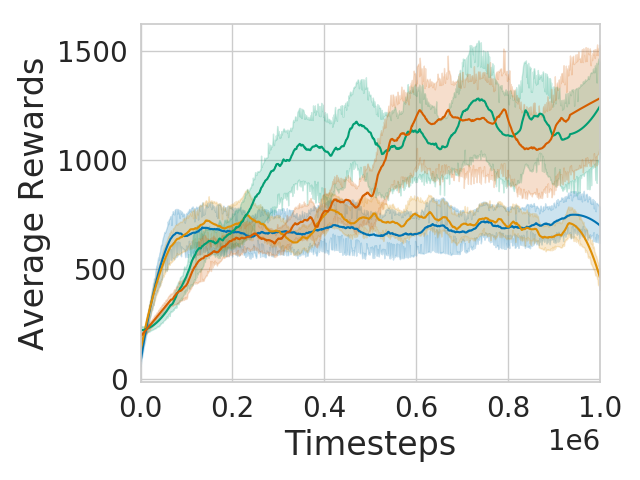}
        \caption{Hopper-v1}
    \end{subfigure}
    \hfill
    \begin{subfigure}{0.3\linewidth}
        \centering
        \includegraphics[width=\linewidth]{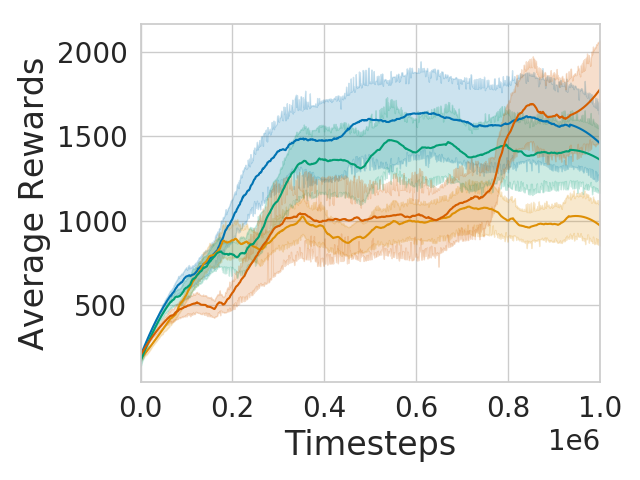}
        \caption{Walker2d-v1}
    \end{subfigure}
    \hfill
    \begin{subfigure}{0.3\linewidth}
        \centering
        \includegraphics[width=\linewidth]{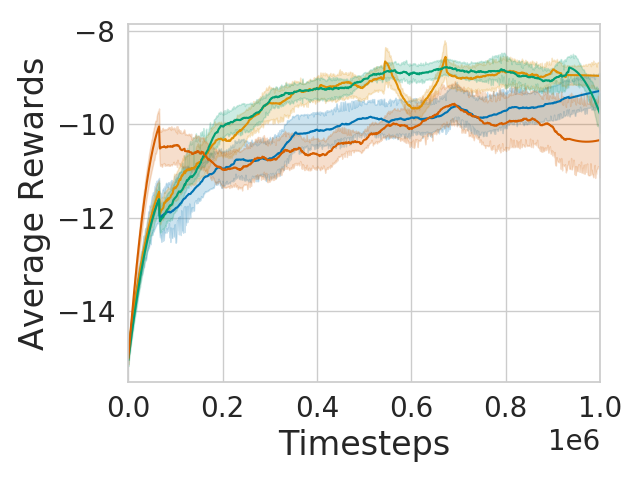}
        \caption{Reacher-v1}
    \end{subfigure}

    \caption{Sample performance curves of TD3 on randomized hyperparameters (see Fig.~\ref{fig:barplots} for full results). Note that TD3 handles hyperparameter perturbations better than DDPG.}
    \label{fig:randomperturbTD3}
\end{figure*}

\begin{figure*}
    \centering
    \begin{subfigure}{0.49\linewidth}
    \centering
    \includegraphics[width=0.8\linewidth]{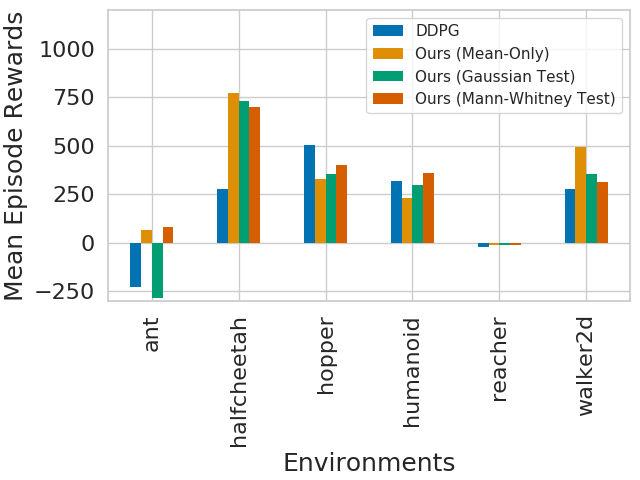}
    \caption{Performance as measured by mean lifetime reward using DDPG as the base algorithm.
    A variant of our recovery method outperforms the baseline in almost all the environments.}
    \end{subfigure}
    \hfill
    \begin{subfigure}{0.49\linewidth}
    \centering
    \includegraphics[width=0.8\linewidth]{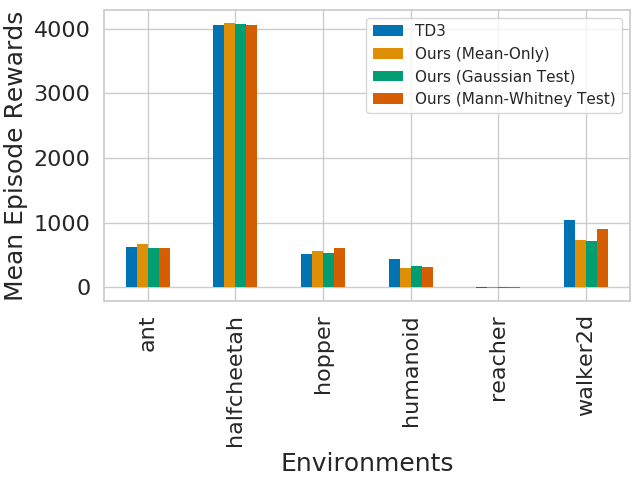}
    \caption{Performance as measured by mean lifetime reward using TD3 as the base algorithm. TD3 is inherently much more stable than DDPG, but our recovery method does not significantly reduce performance}
    \end{subfigure}
    \caption{Performance comparison of all the methods}
    \label{fig:barplots}
\end{figure*}

As shown in Figures \ref{fig:randomperturbDDPG} and \ref{fig:randomperturbTD3}, unaugmented DDPG has a greater tendency to be unstable during training and has difficulty in recovering compared to TD3. Our approach of dynamic monitoring of the policies and reverting to the previously best policy therefore improves DDPG the most, while not significantly affecting the performance of TD3. The particular choice of which variant of our method to use has a small impact, with the Mann-Whitney variant having a greater performance in general (see Figure~\ref{fig:barplots}).

%===============================================================================
\section{Conclusion}

In this paper we address instability resulting from behavior learning approaches that rely on self-collected experience, This unreliable performance and data inefficiency is undesirable in safety-sensitive real-world domains such as robotics. To this end, we develop a principled approach that periodically evaluates the performance improvement of a policy by testing the hypothesis of continual improvement, and reverts to the last checkpoint in the event of a drop in performance. It is also independent of the model, algorithm and, in the case of the Mann-Whitney variant, insensitive to the scale of the reward function, as demonstrated by our experiments using both on- and off-policy algorithms on a variety of simulated robotic continuous control tasks. 

We evaluated our method on a suite of six standard benchmarks for simulated robotic continuous control and compared performance against two state-of-the-art RL algorithms in continuous control, DDPG and TD3. Our results from experiments with high learning rates demonstrated that our method is a viable option for maintaining stability of learning, showing ability to recover from degenerate policies. The experiments with randomly perturbed hyperparameters demonstrated significant improvement over DDPG in 5 out of the 6 evaluation environments and no decrease in performance with TD3, which is known to be relatively stable already. Our proposed approach shows little sensitivity to the precise value of $\rho$ threshold within a reasonable range of values. Given the ability to revert to the best policy, our method can be seen as a form of safe long-term exploration, allowing the agent to explore a variety of directions in parameter space while always being able to recover.

While our method limits performance degradation, in its current state, the policy is not informed by the result of the evaluations. A future version of our method could incorporate the analysis of the evaluation experience, along with the prior training experience, to determine the cause of instability. Furthermore, our approach collects extra data outside of training for evaluation, which may be costly in real-world settings such as robotics. Although this experience is used by the agent during training, reducing wastage, a rolling, online evaluation scheme would eliminate compute cost of our method. Alternately, off-policy evaluations, a developing field focusing on evaluating an agent using a buffer instead of taking steps in the environment \cite{irpan2019off, li2019perspective, xie2019optimal}, are another promising way to reduce evaluation overhead. While this work considered the single-task setting, our approach could be extended to the multi-task setting as an autonomous detector of task transitions in the absence of explicit task labels.

%===============================================================================

\addtolength{\textheight}{-2.9cm}   % This command serves to balance the column lengths

%%%%%%%%%%%%%%%%%%%%%%%%%%%%%%%%%%%%%%%%%%%%%%%%%%%%%%%%%%%%%%%%%%%%%%%%%%%%%%%%
%\section*{ACKNOWLEDGMENTS}

%===============================================================================

\printbibliography

\end{document}